\documentclass[letterpaper, 10 pt, conference]{ieeeconf}  

\IEEEoverridecommandlockouts                              

\usepackage{graphicx} 
\usepackage{float} 
\usepackage{subfig}
\usepackage{algorithm}
\usepackage{array}
\usepackage{algorithmic}

\usepackage{amsfonts}
\usepackage{amsmath,amssymb}
\usepackage{mathrsfs}
\usepackage{cite}
\usepackage{balance}
\usepackage{bbding}
\usepackage{booktabs}
\usepackage[table]{xcolor} 
\usepackage{xcolor}
\usepackage{makecell} 
\usepackage{url}
\usepackage{multirow}
\title{\LARGE \bf
Learning Visuomotor Policy for Multi-Robot Laser Tag Game
}

\author{ 
     Kai Li$^{1,2}$, Shiyu Zhao$^{2}$
      \thanks{ This research work was supported by National Natural Science Foundation of China (Grant No. 62473320). (Corresponding author:Shiyu Zhao.)}
	\thanks{$^{1}$College of Computer Science and Technology at Zhejiang University, Hangzhou, China.}
	\thanks{$^{2}$School of Engineering at Westlake University, Hangzhou, China.
	{\tt\small \{likai,zhaoshiyu\}@westlake.edu.cn}}%
    \thanks{$^{3}$ Code is at this link: https://github.com/woodyhaki/laser-tag }
}

\usepackage[mathscr]{eucal}
\begin{document}

	\maketitle
	\thispagestyle{empty}
	\pagestyle{empty}

	\begin{abstract}
In this paper, we study multi-robot laser tag, a simplified yet practical shooting-game-style task. Classic modular approaches on these tasks face challenges such as limited observability and reliance on depth mapping and inter-robot communication. To overcome these issues, we present an end-to-end visuomotor policy that maps images directly to robot actions. We train a high-performing teacher policy with multi-agent reinforcement learning and distill its knowledge into a vision-based student policy. Technical designs, including a permutation-invariant feature extractor and depth–heatmap input, improve performance over standard architectures. Our policy outperforms classic methods by 16.7\% in hitting accuracy and 6\% in collision avoidance, and is successfully deployed on real robots. Code will be released publicly$^{1}$.
	\end{abstract}

	\section{Introduction}
In this paper, we study multi-robot laser tag using an end-to-end visuomotor policy. The laser tag game originated as a recreational activity in which laser beams are used as simulated bullets to mimic gunfights. It has gained widespread popularity as a form of entertainment, both in physical arenas and in video games, due to its safety, competitive nature, and engaging gameplay. In the robotics community, shooting-based combat games are also used as dynamic competitive platforms to test the robustness of algorithms and system designs. A notable example is the ICRA AI Challenge, which adopted the DJI RoboMaster combat game as its competition platform from 2017 to 2019\cite{robomaster_wiki}. In this autonomous confrontation, two teams of mobile robots engage in battle by firing lasers or projectiles at the opponents. Success requires robots to coordinate strategically, accomplish designated objectives, outmaneuver opponents, and ultimately outperform the rival team in scoring. In this paper, we adopt laser tag as the setting for robot shooting combat.

Beyond its recreational and educational value, the algorithmic components of robot laser tag, such as object detection, enemy state estimation, autonomous aiming, planning, and control, have broad applicability in real-world robotic tasks, such as the capture or interception of malicious drones \cite{zheng2025vision,pliska2024towards}.
\begin{figure}[thpb]
      \centering
      \includegraphics[scale=0.70]{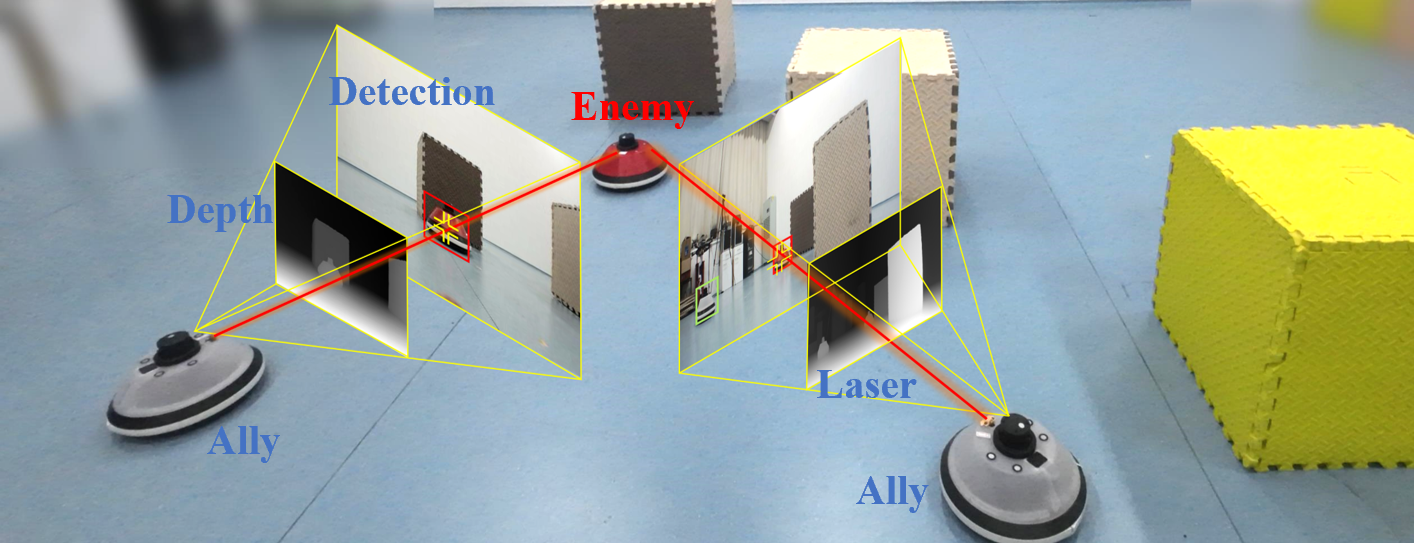}
      \caption{Illustration of the multi-robot laser tag game. The laser beams shown are virtual for visualization purposes.}
      \label{feature_extractor}
\end{figure}
While classic algorithms for shooting-game-style tasks often follow a modular design, they face the following challenges. \textbf{First}, enemy state estimation is a critical component in the classic shooting game pipeline\cite{zheng2025vision,pliska2024towards}. When estimating an adversary's state (e.g. relative position and velocity) from monocular vision, accurate results typically require the robot to follow specific motion patterns that satisfy observability conditions \cite{Ning2024,li2022three}. When these conditions are not met, and combined with the presence of observation noise, the resulting estimates become unreliable, which in turn degrades the overall performance of shooting accuracy.
\textbf{Second}, for obstacle avoidance during the game, some works\cite{chen2025online,huang2024collision} depend on global localization as the policy input, which may not be available in real-world settings. Other works\cite{pliska2024towards,scheidemann2025obstacle,zhou2021ego} regarding similar tasks with onboard sensors require depth-sensor-based mapping for perception and obstacle-avoidance planning, adding the hardware cost and system complexity.
\textbf{Third}, in the multi-robot scenario, previous work\cite{zheng2025vision} relies on inter-robot communication to cooperate to complete the shooting or capture task. This dependency limits the scalability and efficiency of the multi-robot system.

Rethinking from a first-principle's perspective, since humans can play first-person shooting (FPS) games cooperatively, without relying on explicit state estimation, global localization, depth-sensor-based mapping, or mutual communication, \textbf{robots should adopt a similar approach}, relying solely on vision and proprioceptive states to generate actions and accomplish the task. The end-to-end visuomotor policy\cite{Xing2024BootstrappingRL,geles2024demonstrating,zhuang2023robot} directly maps images to actions, which exactly follows the human behavior pattern.

To address the aforementioned challenges, we present a multi-robot laser tag system driven by an end-to-end visuomotor policy. We adopt privileged learning \cite{chen2020learning} to train the policy: a state-based teacher policy is first trained using multi-agent reinforcement learning (MARL), and a vision-based student policy is then distilled from the teacher’s demonstrations. We validate our approach through experiments in both simulation and real-world environments.
The main contributions of this paper are as follows:

1) We present a decentralized, end-to-end visuomotor policy for autonomous multi-robot laser tag game, which operates without explicit state estimation, global localization, wireless communication, or depth-sensor-based mapping.

2) Compared with classic modular approaches for shooting-game-style tasks, our method improves the hit score by up to 16.7\% and reduces the collision rate by 6\%, by eliminating the inaccuracies introduced by enemy state estimation.

3) Technical designs such as the permutation-invariant feature extractor and depth-heatmap input improve performance over standard baseline architectures.

4) We deploy the proposed policy on a real-world multi-robot system with limited onboard computational resources, demonstrating its practicality and effectiveness.
    
\section{Related Works}
Shooting-game-style tasks can be found in many robotic applications such as malicious drone capture and interception\cite{zheng2025vision,pliska2024towards}. Classic modular algorithms for autonomous shooting-game-style tasks include detection, state estimation, planning and control\cite{zheng2025vision,pliska2024towards}. In the RoboMaster competition \cite{robomaster_wiki}, enemy detection and state estimation is commonly achieved by using specially designed visual markers. While effective in the competition setting, this approach has limited applicability, as it relies on artificially designed markers that must be physically attached to the robots.

For markerless enemy detection and state estimation, the common approach is to use image bounding box detections and the corresponding relative position information\cite{Ning2024,vrba2020marker}. To improve estimation accuracy, works such as \cite{Ning2024,li2022three,zheng2025optimal} analyze the motion patterns required for the pursuer to make the enemy's state observable from monocular cameras, in both single\cite{Ning2024,li2022three} and multi-robot\cite{zheng2025optimal} settings. However, in practice, the adversarial behavior of the enemy makes these conditions difficult to satisfy.
For the planning and control module, some works\cite{chen2025online,huang2024collision} adopts MARL and rely on global localization as the input of the policy to pursue enemies while avoiding obstacles. However, these methods become impractical when global localization is unavailable, for example under motion tracking system or satellite signal denial. The work in \cite{bajcsy2024learning} employs onboard sensing to pursue enemies in open space, but does not account for obstacle avoidance. Moreover, it depends on depth sensors for enemy detection and state estimation. Other works \cite{pliska2024towards,scheidemann2025obstacle,YOPO,zhou2021ego} leverage depth sensors for perception, mapping or planning. Notably, the work in \cite{su2025toward} employs MARL with LiDAR observations for physical robot adversarial games. While effective, depth sensors substantially increase system complexity and hardware cost, posing significant challenges for low-cost and lightweight robotic platforms. 

To overcome the limitations of classical modular pipelines in shooting-game-style tasks, we adopt a monocular visuomotor policy for laser tag games, enabling end-to-end decision making without relying on depth-based environmental mapping, explicit state estimation, or trajectory planning. Distinct from prior MARL and onboard end-to-end approaches, our method targets fully decentralized adversarial interactions under partial observability, using only onboard monocular vision, which improves system simplicity and real-world deployability.

\begin{figure*}[thpb]
      \centering
      \includegraphics[scale=0.48]{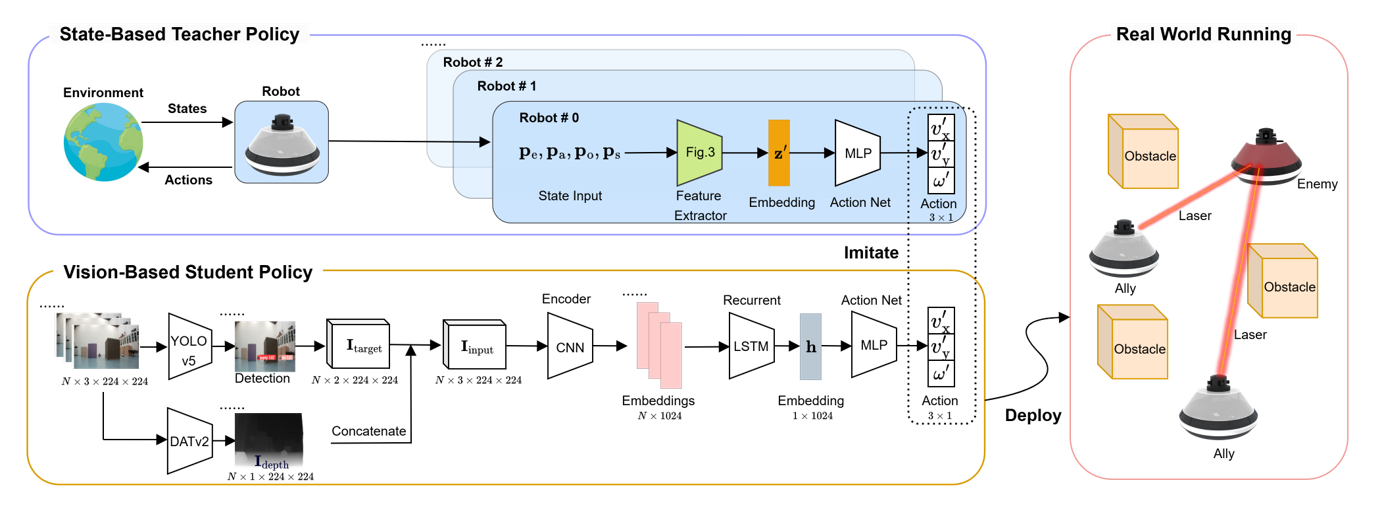}
      \caption{This figure shows the pipeline of our method. The teacher policy takes privileged states and outputs velocity command, and is trained with MARL. The student policy takes a time series of images and outputs velocity command. The student policy imitates the action output of the teacher and is deployed onboard. Shape of each tensor in the policy is shown. $N$ denotes the number of historical images used for the recurrent module. DATv2 is the monocular depth estimation method Depth Anything v2\cite{depth_anything_v2}. }
      \label{pipeline}
\end{figure*}
\section{Proposed Method}
\subsection{Problem Formulation}
The goal of this work is to learn a policy $\pi$ that directly maps image observations to control commands, and deployed on a decentralized multi-robot system to complete the laser tag game. The policy $\pi$ is formulated as $\mathbf{a}^{t}=\pi(\mathbf{h}^{t})$,  where the observation input $\mathbf{h}^{t}$ at time $t$ is the embedding from a time series of \( N \) image observations, and the action $\mathbf{a}^{t}$ is the robot's velocity command.
For the laser tag game, the robots are divided into two teams. 
We refer to the robots in Team~I as \emph{allies} and those in Team~II as \emph{enemies}, 
and use these terms consistently throughout the paper. 
The objective of Team~I (allies) is to locate and shoot the robots of Team~II (enemies) using onboard laser emitters, 
while the objective of Team~II is to escape from the attacks of Team~I. 
Robots from both teams must avoid collisions with obstacles and other robots, 
and allies must also avoid accidentally shooting teammates. 
Robots in Team I and Team II are adversary and employ adversary policies.

\begin{figure}[thpb]
      \centering
      \includegraphics[scale=0.25]{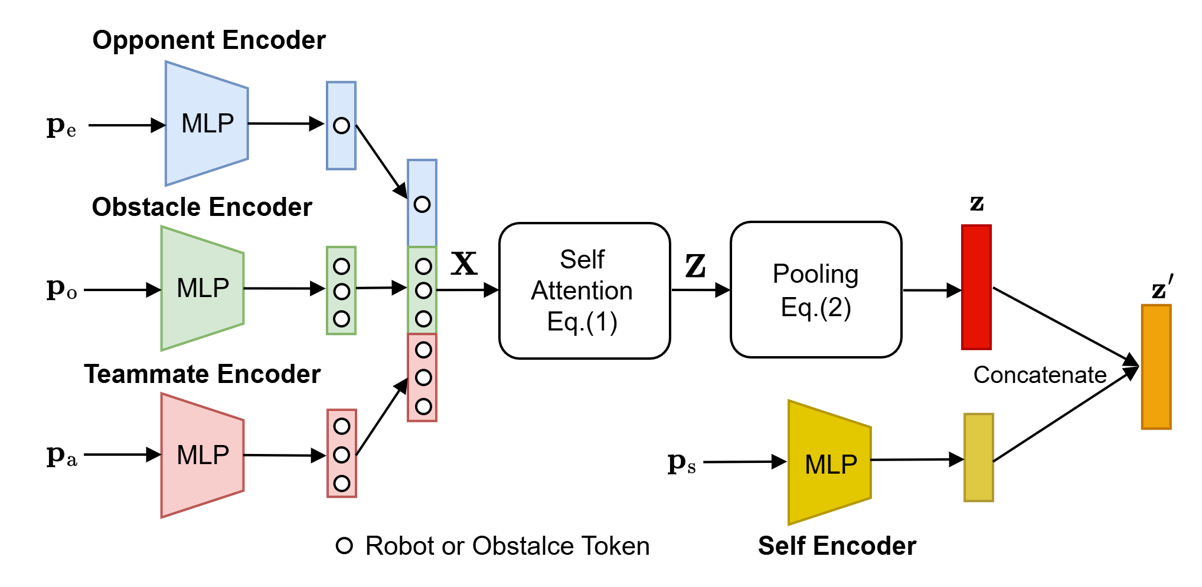}
      \caption{Structure of the feature extractor of the teacher policy. Opponent, obstacle and teammate states are encoded with self-attention and summation pooling. Each entity token (White circle) in the figure represents one embedded instance of the obstacle or neighbor robot. 
      }
      \label{feature_extractor}
\end{figure}
\subsection{Overview}
To learn the visuomotor policy for the laser tag game, we choose the privileged imitation learning\cite{chen2020learning}. The privileged imitation learning first learns a state-based teacher policy with privileged states.
Then, the state-based teacher policy is distilled into a vision-based student policy. Here, state-based means the inputs of the teacher policy is the state information, such as the relative position of the obstacles or adversary robots. 
The motivations of this design choice are as follows. Direct vision-based reinforcement learning suffers from low sample efficiency due to the high-dimensional pixel space and its trial-and-error nature. In contrast, vision-based imitation learning is popular for visuomotor policies due to its higher sample efficiency and straightforward form. The privileged-state-based teacher policy can serve as a data generator for imitation learning.
Since our task is a multi-robot laser tag game, and reinforcement learning has demonstrated impressive results in robot games like drone racing\cite{Xing2024BootstrappingRL,geles2024demonstrating}, we use MARL with privileged states to train the teacher policy.
\subsection{Teacher Policy via Multi-Agent Reinforcement Learning} 

\textbf{State and Action Space.} 
The teacher policy maps state inputs to control commands.
The state of the $i$-th robot at time $t$ is defined as
$
\mathbf{s}_i^t = [\mathbf{p}_{\rm{e}}^t, \mathbf{p}_{\rm{a}}^t, \mathbf{p}_{\rm{o}}^t, \mathbf{p}_{\rm{s}}^t],
$
where $\mathbf{p}_{\rm{o}}^t$ denotes the relative positions of the $K$ nearest obstacles, 
$\mathbf{p}_{\rm{a}}^t$ the relative positions of the $K$ nearest teammates, 
$\mathbf{p}_{\rm{e}}^t$ the relative position of the opponents, 
and $\mathbf{p}_{\rm{s}}^t = [x_{\rm{s}}^t, y_{\rm{s}}^t, \psi_{\rm{s}}^t]$ the robot’s own state (global 2D position and heading). 
All relative positions are expressed in the local coordinate frame of the robot. 
To ensure a fixed input dimension regardless of the number of teammates or obstacles, 
we adopt a $K$-nearest selection strategy to construct the state vector.
The action space is a 3-dimensional vector,
$
\mathbf{a}_i^t = [v_x^t, v_y^t, \omega^t],
$
where $[v_x^t, v_y^t]$ is the linear velocity and $\omega^t$ is the angular velocity.

\textbf{Feature Extractor}.
As shown in Fig.~\ref{pipeline}, the raw state is embedded by a feature extractor before being sent into the action network.
Fig.~\ref{feature_extractor} shows the architecture of the feature extractor. Notably, the state/action space and the feature extractor are agnostic to whether a robot is an ally or an enemy. We choose self-attention\cite{vaswani2017attention} as the core component for the feature extractor, since it has shown improved performance over other forms of feature extractors in MARL\cite{chen2025online,huang2024collision}, due to its strong representation power of mutual relation. Each individual part of the raw state is first encoded by a multi-layer perceptron (MLP), and the resulting embeddings are concatenated and passed through a self-attention layer as the key, query and value. The output of the self-attention layer is,
\begin{equation}
\mathbf{Z} = \mathrm{Attention}(\mathbf{X}, \mathbf{X}, \mathbf{X}) \in \mathbb{R}^{B \times M \times F},
\end{equation}
where $\mathbf{X}$ is the concatenated input embeddings, $B$ is the batch size, $M$ is the number of embedded obstacle or robot tokens, and $F$ is the feature dimension.
To obtain a fused embedding, a weighted summation pooling is applied along the embedded token axis:
\begin{equation}
\mathbf{z} = \text{MLP}(\sum_{i=1}^{M} \alpha_i \cdot \mathbf{Z}_i) \in \mathbb{R}^{B \times F^{\prime}},
\end{equation}
where $F^{\prime}$ is the fused embedding dimension. The attention weights $\alpha_i$ are computed by applying softmax normalization to a score $s_i$ derived from the mean of each attention embedding $\mathbf{Z}_i$,
\begin{equation}
s_i = \frac{1}{F} \sum_{j=1}^{F} Z_{i,j}, \quad \alpha_i = \frac{\exp(s_i)}{\sum_{k=1}^{M} \exp(s_k)}.
\end{equation}
The fused embedding $\mathbf{z}$ is concatenated with the embedding of self state $\mathbf{p}_{s}$ and produces the final embedding $\mathbf{z}^{\prime}$.  $\mathbf{z}^{\prime}$ is used as the input of the actor and critic network.
The motivation for the feature extractor design is as follows. Since we adopt a $K$-nearest strategy for neighbor selection, the chosen neighbors and their orders may change abruptly in dynamic environments, introducing discontinuities in the state input. Such discontinuities can lead to unstable outputs of the actor network and oscillatory robot behavior, particularly under real-world deployment. The attention-based summation pooling feature is \textit{permutation-invariant}. A permutation-invariant feature extractor aggregates neighbor information as a set and ensures that the embedding $\mathbf{z}$ is independent of input order, thereby yielding smoother and more robust policy behavior in dynamic multi-robot-multi-obstacle environments. Furthermore, in multi-robot systems, a robot is agnostic to the identities and ordering of its neighbors; therefore, it is desirable for the learned features to remain invariant to input ordering. 

\textbf{Reward Function}.
Since we have two teams of robots acting in adversary form, the reward functions are also in different forms. For Team~I (ally), the team of robots that actively attack enemies, the reward function at time $t$ is formulated as $r^{t1}=r_{\rm{dist}}^{t1} + r_{\rm{aim}}^{t1} + r_{\rm{obst}}^{t1} + r_{\rm{bound}}^{t1} + r_{\rm{cf}}^{t1}$, 
where $r_{\rm{dist}}^{t1}$ denotes the distance reward with respect to the enemy robot, $r_{\rm{aim}}^{t1}$ denotes the laser aiming reward with respect to the enemy robot, $r_{\rm{obst}}^{t1}$ denotes the obstacle avoidance reward, $r_{\rm{bound}}^{t1}$ denotes the boundary avoidance reward and $r_{\rm{cf}}^{t1}$ denotes the cross fire avoidance reward. Each reward components of Team I are formulated as,
\begin{equation}  
\begin{split}    
r_{\rm{dist}}^{t1} &= -\lambda_{0}\, \lvert d - d_{\rm{desi}} \rvert \\
r_{\rm{aim}}^{t1}  &= -\lambda_{1}\, \lvert \theta_{\rm{i}}-\theta_{\rm{e}} \rvert\\    
r_{\rm{obst}}^{t1} &= -10 \ \text{if hits the obstacles} \\  
r_{\rm{bound}}^{t1} &= -10 \ \text{if hits the boundaries}  \\
r_{\rm{cf}}^{t1} &= -10 \ \text{if ally robots enter the shooting zone},  
\end{split}
\label{reward_componet_ally}
\end{equation}
where $d$ is the distance to the enemy, $d_{\rm{desi}}$ the desired shooting distance, $\theta_{\rm{i}}$ the heading angle of the laser emitter in local frame, and $\theta_{\rm{e}}$ the heading angle of the enemy relative to the ally. This reward formulation encourages the ally robots to aim at enemy robots while avoiding obstacles and accidental attack on allies. For Team~II (enemy), the reward function is formulated as $r^{t2}=r_{\rm{dist}}^{t2} + r_{\rm{aim}}^{t2} + r_{\rm{obst}}^{t2} + r_{\rm{bound}}^{t2}$. Each reward components of Team II are formulated as:
\begin{equation}  
\begin{split}    
r_{\rm{dist}}^{t2} &= \lambda_{2}\ d \\
r_{\rm{aim}}^{t2}  &= \lambda_{3}\, \lvert \theta_{\rm{i}}-\theta_{\rm{e}} \rvert\\    
r_{\rm{obst}}^{t2} &= -10 \ \text{if hits the obstacles} \\  
r_{\rm{bound}}^{t2} &= -10 \ \text{if hits the boundaries}.
\end{split}
\label{reward_componet_ally}
\end{equation}
$\lambda_{0}$ to $\lambda_{3}$ are different hyperparameters. This reward formulation encourages the enemy robot in Team~II to avoid being shot while avoiding obstacles.

\textbf{MARL Algorithm}. We use MADDPG \cite{lowe2017multi}, which follows the centralized training with decentralized execution paradigm, to train the adversary policies for both teams.

\subsection{Student Policy via Vision-Based Imitation Learning} 
\begin{figure}[thpb]
      \centering
      \includegraphics[scale=0.21]{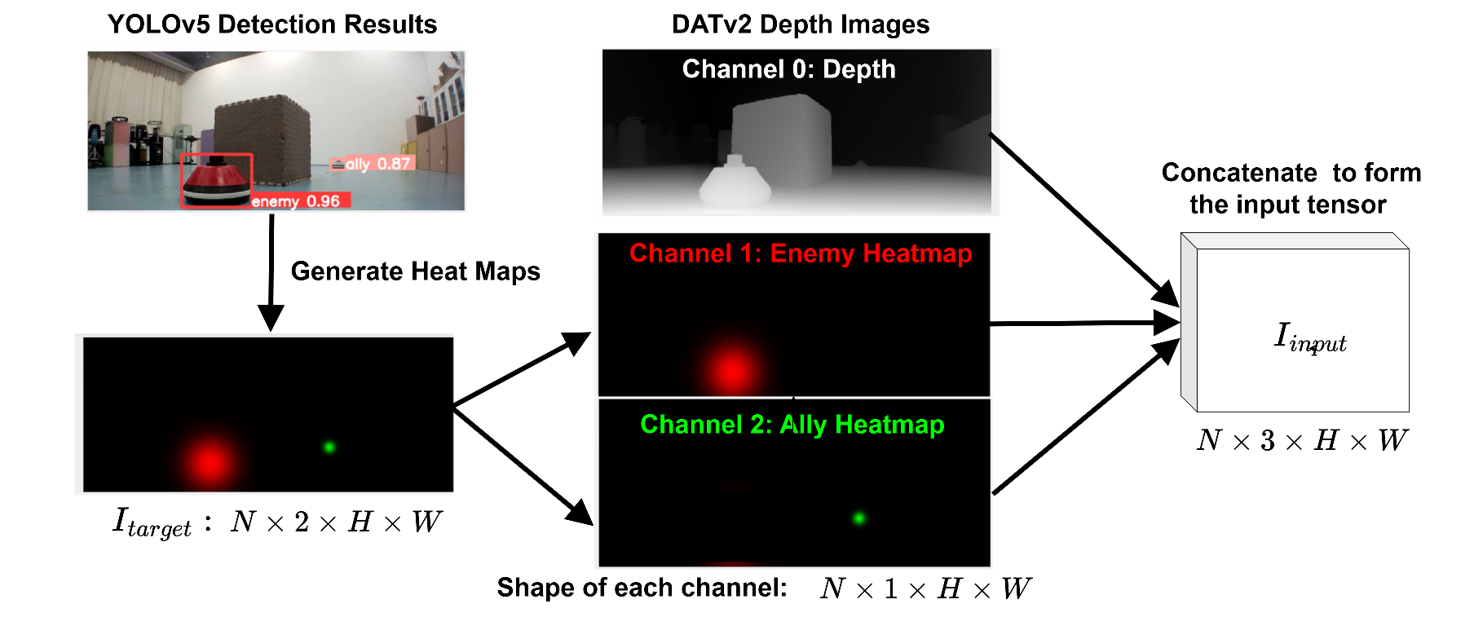}
      \caption{ The original image is sent into YOLOv5 and DATv2 to generate detection results and depth images. The detection results contain two classes, namely the enemy and the ally. A heat map that indicates the detection location is generated for each class. Then the two classes of heat maps are concatenated with the depth image along the channel axis to form the input tensor.   }
      \label{visuo_input}
 \end{figure}

As shown in Fig.~\ref{pipeline}, the student policy takes a time series of $N$ onboard camera images and outputs the velocity command. The input image of each time stamp is processed as follows. First, we use a YOLOv5 Nano\cite{redmon2016you} object detector to detect the enemy and ally robots. Then a Gaussian kernel is placed in the detected bounding box area to create a target area image $\mathbf{I}_{\rm{target}}$. The target area image serves as an indicator for the enemy and ally location in the image space. Meanwhile, the original RGB image is fed into DepthAnythingV2~\cite{depth_anything_v2} (DATv2) to generate a depth image $\mathbf{I}_{\rm{depth}}$. The depth values are clipped to the 5th and 95th percentiles and normalized to the range $[0,1]$. Finally, the target image $\mathbf{I}_{\rm{target}}$ and the depth image $\mathbf{I}_{\rm{depth}}$ are concatenated along the channel axis to form the input tensor $\mathbf{I}_{\rm{input}}$. While previous works use image edges\cite{geles2024demonstrating} or raw images\cite{Xing2024BootstrappingRL} as the input, here we convert the bounding box detection to a Gaussian kernel. The reason for this conversion is that our goal is to make the laser emitter to aim at the enemy robot. Therefore, we want to make the aiming as accurate as possible. The shape of the Gaussian kernel indicates a higher weight in the center and a lower weight in the boundary area, which reflects the aiming goal. The Gaussian blur kernel size $s$ for the heatmap is defined as
$
s = \lambda/{D},
$
where $D$ denotes the relative distance to the detected robot and $\lambda$ is a scaling factor. 
Thus, closer robots correspond to larger kernel sizes in the heatmap. \textbf{Notably, $D$ is estimated from the known physical size of the robot, bounding box width and the camera focal length, rather than from the non-metric depth output of DATv2}. The depth maps are appearance-invariant compared with monocular images, which enhances the robustness of the visuomotor policy. Fig.~\ref{visuo_input} shows how $\mathbf{I}_{\rm{input}}$ is formulated.
 
The input tensor $\mathbf{I}_{\rm{input}}$ is sent into an image encoder of convolutional neural network (CNN) to be encoded as an embedding. The time series of image embeddings are then sent into a long short-term memory (LSTM) recurrent module to be further encoded as a fused embedding $\mathbf{h}^t$. Finally, $\mathbf{h}^t$ is sent to an MLP to regress the action. Since the onboard camera has a limited field-of-view (FOV), we choose to feed a time series of images and use LSTM to incorporate temporal information to increase the volume of information for decision-making. The LSTM also mitigates the temporal inconsistency of the depth estimates produced by DATv2 by aggregating information across time.
The imitation learning loss is defined as the squared L2 norm between the student action and the teacher action,
\begin{equation}  
\begin{split}    
\mathcal{L} =  \| \pi_{\rm{stu},\theta}(\mathbf{h}^{t}) - \pi_{\rm{tea}}(\mathbf{s}^{t})\|_{2},
\end{split}
\label{reward_componet_ally}
\end{equation}
where $\theta$ denotes the learnable parameters of the student. Both the image encoder and the following action part are trained end-to-end and no part of the module is frozen.
\section{Experimental Evaluations}
\subsection{Implementation Details and Policy Training}
\textbf{Teacher Policy Training}. We train state-based adversary agents using MADDPG \cite{lowe2017multi}. The actor is a two-layer MLP (128 units), and the critic is a three-layer MLP (512 units).
MLPs in the feature extractor are two-layer with 128 hidden units.
Training uses a batch size of 512, a replay buffer of 10,000, and learning rates of 1e-3 for both networks. The feature extractor (with $K=3$ nearest neighbors) shares the actor’s optimizer and learning rate. The setting involves 1 to 3 ally robots against a single enemy robot.

\begin{table*}[h]
\setlength\tabcolsep{4.8pt}
\caption{Comparison and ablation of visual encoders and input types, evaluated by Action Error (AE), Collision Rate (CR), and Hit@15. SimpleCNN refers to a lightweight CNN with channel sizes [16, 32, 32] and filter sizes [5, 4, 3]. Onboard inference time (mean with 99\% percentile in parentheses) is reported only for the visual encoder. “train” means in the same training environment, and “test” means in unseen test environments.
We compare RGB, edge images \cite{geles2024demonstrating}, depth, and our input (depth + heatmaps, see Fig.~\ref{visuo_input}).
}
\begin{center}
\begin{tabular}{p{2cm} l c c c c c c c c c c c c p{2cm}}
\toprule
\multirow{2}{*}{\makecell[c]{\textbf{Visual}\\\textbf{Encoders}}}  
& \multirow{2}{*}{
\makecell[c]{\textbf{Input}\\\textbf{Types}}} 
&\multicolumn{3}{c}{\textbf{Simulation Train}} & \multicolumn{3}{c}{\textbf{Simulation Test}} & \multicolumn{3}{c}{\textbf{Real-World Test}}  &  \multirow{2}{*}{\textbf{Time (ms)}}\\
                \cmidrule(lr){3-5} \cmidrule(lr){6-8} \cmidrule(lr){9-11} & 
                                                                        & AE$\downarrow$   & Hit@15 $\uparrow$& CR[\%]$\downarrow$ &   AE$\downarrow$   &  Hit@15$\uparrow$  & CR [\%]$\downarrow$   &  AE$\downarrow$   & Hit@15$\uparrow$  &  CR [\%]$\downarrow$       \\
\midrule
ResNet50\cite{he2016deep}           & RGB                                &  0.0012        &  25.8 ± 6.4         & 40.0   & 0.0075   & 14.3 ± 4.0   &  66.0  &  0.0115  & 0.0 ± 0.0         & 70.0  &  21.6 (28.1)  \\
CLIP RNet\cite{radford2021learning} & RGB                                &  0.0013        &  21.8 ± 3.0         & 54.0   & 0.0066   & 6.4 ± 2.4    &  58.0  &  0.0102  & 0.0 ± 0.0         & 54.0  &  20.1 (25.4)  \\
DINOv2\cite{oquab2024dinov2}        & RGB                                &  0.0020        &  17.4 ± 3.0         & 30.0   & 0.0078   & 6.1 ± 2.6    &  40.0  &  0.0228  & 0.0 ± 0.0         & 50.0  &  137.2 (185.7)\\
MAE\cite{he2022masked}              & RGB                                &  0.0009        &  33.8 ± 1.8         & 38.0   & 0.0015   & 19.7 ± 3.0   &  42.0  &  0.0099  & 7.8 ± 5.2         & 44.0  &  24.9 (28.0)  \\
SimpleCNN                           & Edges\cite{geles2024demonstrating} &  0.0006        &  26.8 ± 2.8         & 38.0   & 0.0020   & 20.8 ± 4.1   &  46.0  &  0.0032  & 12.5 ± 5.9        & 72.0  &  \textbf{7.8 (18.2)}  \\
ResNet18\cite{he2016deep}           & Edges\cite{geles2024demonstrating} &  0.0011        &  26.4 ± 2.9         & 42.0   & 0.0026   & 20.3 ± 4.2   &  40.0  &  0.0050  & 18.8 ± 3.0        & 60.0  &  16.7 (25.6)  \\
SimpleCNN                           & Depth                              &  0.0014        &  20.1 ± 5.0         & 32.0   & 0.0017   & 11.5 ± 2.6   &  40.0  &  0.0030  & 10.1 ± 3.9        & 38.0  &  \textbf{7.8 (18.2)}  \\
ResNet18\cite{he2016deep}           & Depth                              & \textbf{0.0005}&  22.0 ± 2.4         & 30.0   & 0.0015   & 14.3 ± 2.2   &  36.0  &  0.0029  & 12.5 ± 3.7        & 38.0  &  16.7 (25.6)  \\\rowcolor{gray!20} 
SimpleCNN                           & \textbf{Ours}                      & 0.0006         &  36.0 ± 1.7         & 32.0   & 0.0010   & 33.0 ± 4.1   &  38.0  &  0.0014  & 29.9 ± 5.4        & 40.0  &  \textbf{7.8 (18.2)}   \\\rowcolor{gray!20} 
ResNet18\cite{he2016deep}           & \textbf{Ours}                      & 0.0006         & \textbf{41.6 ± 2.2} & \textbf{28.0}   & \textbf{0.0009}   & \textbf{38.9 ± 2.9} &  \textbf{36.0}  &  \textbf{0.0012}  & \textbf{31.0 ± 5.5} & \textbf{36.0}  &  16.7 (25.6)  \\
\bottomrule
\end{tabular}
\end{center}
\label{table_big_vision}
\end{table*}
\textbf{Student Policy Training}.
The vision-based student policy is trained using imitation learning with dataset aggregation (DAgger) \cite{ross2011reduction}. To facilitate rapid prototyping, we create a digital twin of our multi-robot system in the ROS Gazebo\cite{koenig2004design}. 
The training data are collected from both the simulation and the real-world environment. 4 different simulation scenes are used for data collection, which enriches the data distribution. In total, We collected 80k images from the simulator and 20k from the real world, generated from rollouts of the state-based teacher policy.
The one-layer LSTM contains 128 hidden units, and the action net is a two-layer MLP with 128 hidden units. 
The learning rate of the visual encoder is set at 1e-06, while the learning rate for the action part is set at 1e-05. The memory sequence length for the LSTM is 5. The dimension of the embedding $\mathbf{h}$ is 1024.
The policy is trained on an RTX 4090 GPU for 100 epochs. 

\textbf{Real-World Robot System}.
To evaluate our method in the real world, we construct a multi-robot system in which each robot supports omnidirectional mobility and shares an identical mechanical and sensor configuration. The onboard computer is an Nvidia Jetson Orin NX, featuring 6 ARM cores, 16 GB memory, 25 W power consumption, and up to 100 TFLOPS of computing performance. To distinguish between ally and enemy robots, enemy robots are colored red, while ally robots are colored white. A laser emitter of 5 mW is installed on the ally robot and manually calibrated with the optical axis of the camera. The front-facing camera has a 110° horizontal FOV and a 30 Hz frame rate.

\textbf{Experiment Configurations}. The experiment environment is a 4m $\times$ 4m area. We randomly put obstacles of different shapes in the environment, and the obstacle density (obstacle area / total area) varies from 0 to 0.18. The number of obstacles varies from 0 to 8. 
Each robot is initialized with a random position and heading angle, while the enemy is constrained to lie within the ally’s camera FOV at the start. The ally fires the laser when the enemy is within 50 pixels of the crosshair center and the bounding box width exceeds 20 pixels (i.e., when the physical distance is less than 3 m). The robot detection distance $D$ is from 0.5 m to 3 m. 
Across multiple experiments, the robot’s linear velocity limit is varied from 0.05 m/s to 0.40 m/s, with the angular velocity capped at 2.5 rad/s. 
The speed limits for the ally and the enemy are set the same for a fair game. In the experiment, the enemy uses the state-based adversary policy trained in the teacher-training stage. The state information for the enemy is provided by the motion tracking system. By training on mixed simulated and real-world data and using DATv2 to generate depth images in both domains, the sim-to-real gap is inherently reduced, without requiring additional adaptation methods. We adopt the “Large” version of DATv2 for non-metric depth estimation, which contains 335.3 M parameters. The full size of the student model is 15.9 M with Resnet18.

    \begin{figure*}[thpb]
      \centering
      \includegraphics[scale=0.18]{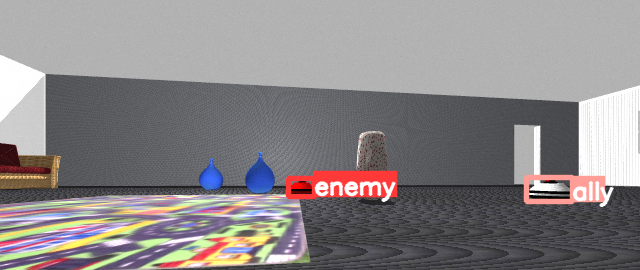}
      \includegraphics[scale=0.18]{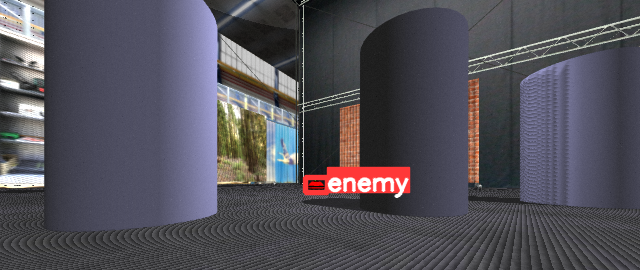}
      \includegraphics[scale=0.18]{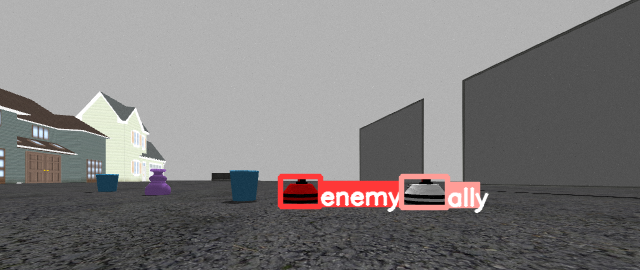}
      \includegraphics[scale=0.18]{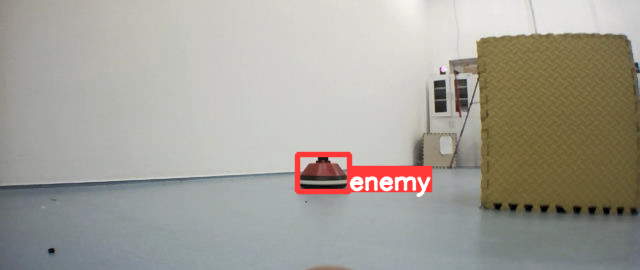}
      
      \includegraphics[scale=0.18]{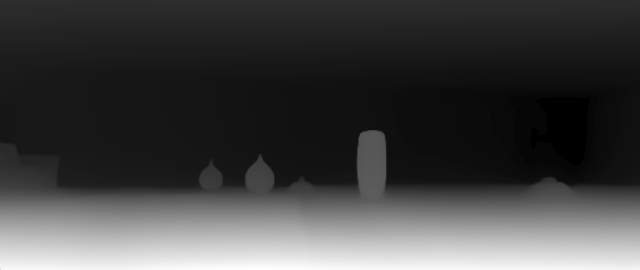}     
      \includegraphics[scale=0.18]{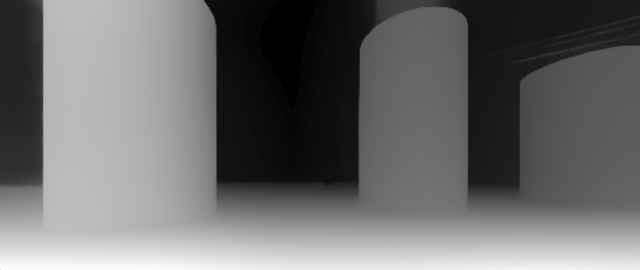}     
      \includegraphics[scale=0.18]{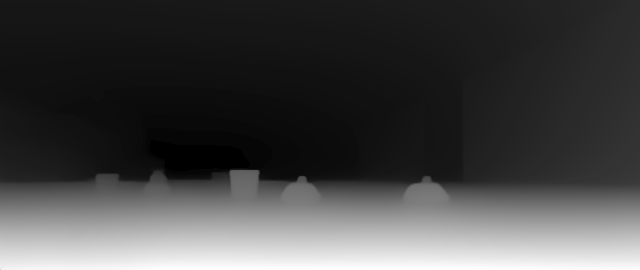}     
      \includegraphics[scale=0.18]{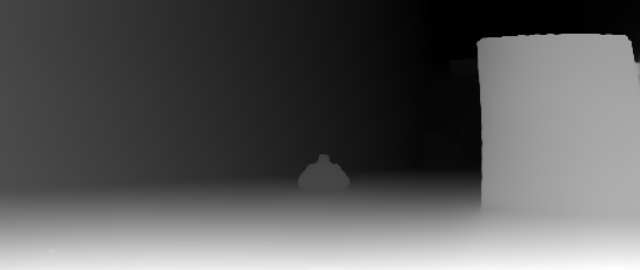}     

      \includegraphics[scale=0.18]{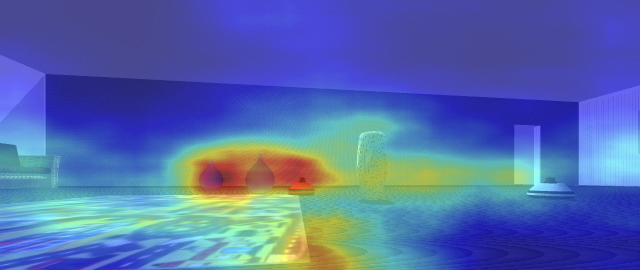}     
      \includegraphics[scale=0.18]{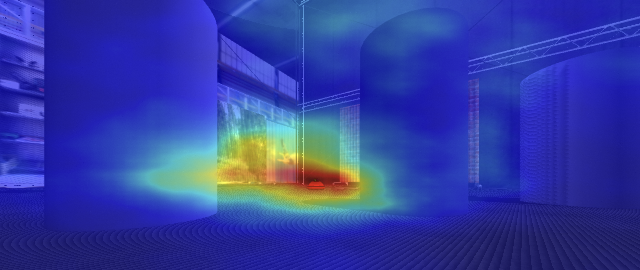}     
      \includegraphics[scale=0.18]{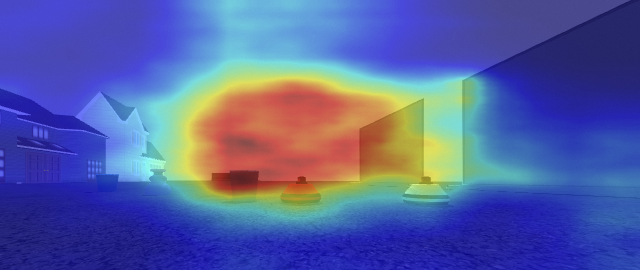}     
      \includegraphics[scale=0.18]{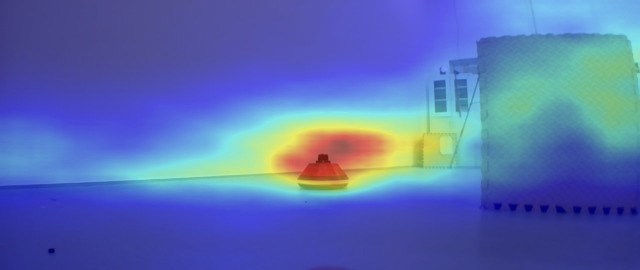}     
        
      \caption{The first row shows enemy and ally detection results, the second row shows depth maps, and the third row shows full-gradient attention maps \cite{srinivas2019full}. The attention maps highlight task-relevant regions of the input image, typically focusing on enemy targets and obstacles, consistent with human intuition. The first three columns present simulation data, and the last column presents real-world data.}
      \label{figure_detection}
 \end{figure*}

\subsection{Results of the Policy}
We first conduct a comparative and ablation study to quantitatively evaluate our visuomotor policy, reporting results across different combinations of visual encoders and input types. We repeat each policy for 100 runs in simulation and 50 runs in the real-world. For all experiments, each episode lasts 2000 image frames (around 70 s).
Performance is measured using several key metrics:
\begin{itemize}
    \item \textbf{Action Error (AE)}: The action discrepancy between the student and the teacher, directly measuring the quality of imitation learning.
    \item \textbf{Hit@15} and \textbf{Hit@30}: The number of times an ally robot hits the enemy for 15 or 30 consecutive frames in one episode. Each value is averaged over 50 trials.
    \item \textbf{Collision Rate (CR)}: The ratio of trials in which an ally collides with obstacles.
\end{itemize}
For the experimental setup, encoders with RGB inputs use pretrained weights without finetuning. In contrast, encoders with edge or depth-included inputs are trained end-to-end with the action network.
Table~\ref{table_big_vision} shows that our depth-plus-heatmap input achieves the best overall performance. When the heatmaps are ablated (depth-only input), hitting performance degrades due to the lack of salient enemy and ally cues. 
Although high-capacity pre-trained encoders such as DINOv2 \cite{oquab2024dinov2} and CLIP \cite{radford2021learning} offer strong general-purpose visual representations, they yield larger action errors and lower hit rates on our test scenes, likely because they are not tailored to the specific robot task. Pure RGB or edge-based inputs \cite{geles2024demonstrating} achieve lower performance, especially in unseen test environments. In contrast, depth images combined target heatmaps, being largely appearance-invariant, facilitate the student in learning correct behaviors across different scenes.
All visual encoders, as well as DATv2 and YOLOv5, are converted to half-precision (FP16) TensorRT models for onboard acceleration. We measure the inference time for each model, all of which process images of size $3\times224\times224$ pixels.
The DATv2 module alone takes 20.2 ± 5.1 ms, and the YOLOv5 Nano takes 5.15 ± 1.2 ms. With all visual modules running in parallel processes, the control frequency of our approach reaches 20~Hz after considering the additional latency from DATv2 and YOLOv5. This ensures smooth robot movements during the game. High-capacity encoders such as DINOv2 cannot run fast enough onboard (more than 100 ms inference time), which limits their practical usage.  
\begin{table}[h]
\setlength\tabcolsep{5pt}
\caption{The gray rows are results of different teacher policies, and the white are those of the students. 
Teacher-SAP refers to the teacher policy with our self-attention-and-pooling as the feature extractor. Teacher-SA refers to policy with vanilla self-attention feature extractor used in MARL\cite{chen2025online} without permutation-invariant operators. Teacher-SA+SDF refers to \cite{huang2024collision}'s feature extractor design. BC denotes behavioral cloning. }
\begin{center}
\begin{tabular}{l c c  c c}
\toprule
\textbf{Action Methods}    &  \textbf{AE}$\downarrow$     & \textbf{Hit@15}$\uparrow$ &   \textbf{Hit@30}$\uparrow$  &  \textbf{CR [\%]}$\downarrow$  \\\rowcolor{gray!20} 
\midrule  
Teacher-MLP                             & N/A    &  40.6 ± 1.6  &  19.0 ± 1.7  &  \textbf{6.0}                      \\ \rowcolor{gray!20} 
Teacher-SA+SDF\cite{huang2024collision} & N/A    &  40.9 ± 2.4  &  17.6 ± 2.0  &  \textbf{6.0}                      \\ \rowcolor{gray!20} 
Teacher-SA\cite{chen2025online}         & N/A    &  42.3 ± 2.0  &  18.0 ± 2.9  &   10.0                             \\ \rowcolor{gray!20} 
\textbf{Teacher-SAP (Ours)}             & N/A    &  \textbf{48.6 ± 1.7}  &  \textbf{22.1 ± 2.0}      &     10.0                     \\ 

BC-MLP                                  & 0.0251 &  0.0   ± 0.0        & 0.0 ± 0.0       &     80.0                     \\
BC-LSTM                                 & 0.0059 &  5.3 ± 1.7        & 0.0 ± 0.0       &     48.0                     \\
DAgger-MLP                              & 0.0010 &  29.6 ± 2.5       & 12.4 ± 3.2    &     40.0                     \\
DAgger-LSTM                             & 0.0006 &  38.4 ± 2.2       & 17.2 ± 2.8    &     36.0                     \\

\bottomrule
\end{tabular}
\end{center}
\label{table_motor}
\end{table}

Next, we compare the performance of the teacher policy trained with different feature extractors in Table~\ref{table_motor}. Using self-attention with pooling improves laser-hitting performance over a plain MLP or a self-attention extractor alone. When the permutation-invariant pooling is ablated, corresponding to the feature design of \cite{chen2025online}, the policy’s performance degrades. This degradation arises because, without permutation-invariant features, the policy tends to produce oscillatory robot movements in practice, causing the robot to get stuck and preventing smooth chasing and aiming at the enemy.
We also conduct a comparative study to evaluate performance in terms of different imitation learning methods and action network choices.
For the student policy, the DAgger-based approach combined with an LSTM achieves the best performance. This can be attributed to the robot’s limited camera FOV, where the LSTM’s capacity to aggregate temporal information enhances the robustness of decision-making. 
\begin{figure*}[thpb]
      \centering
      \includegraphics[scale=0.91]{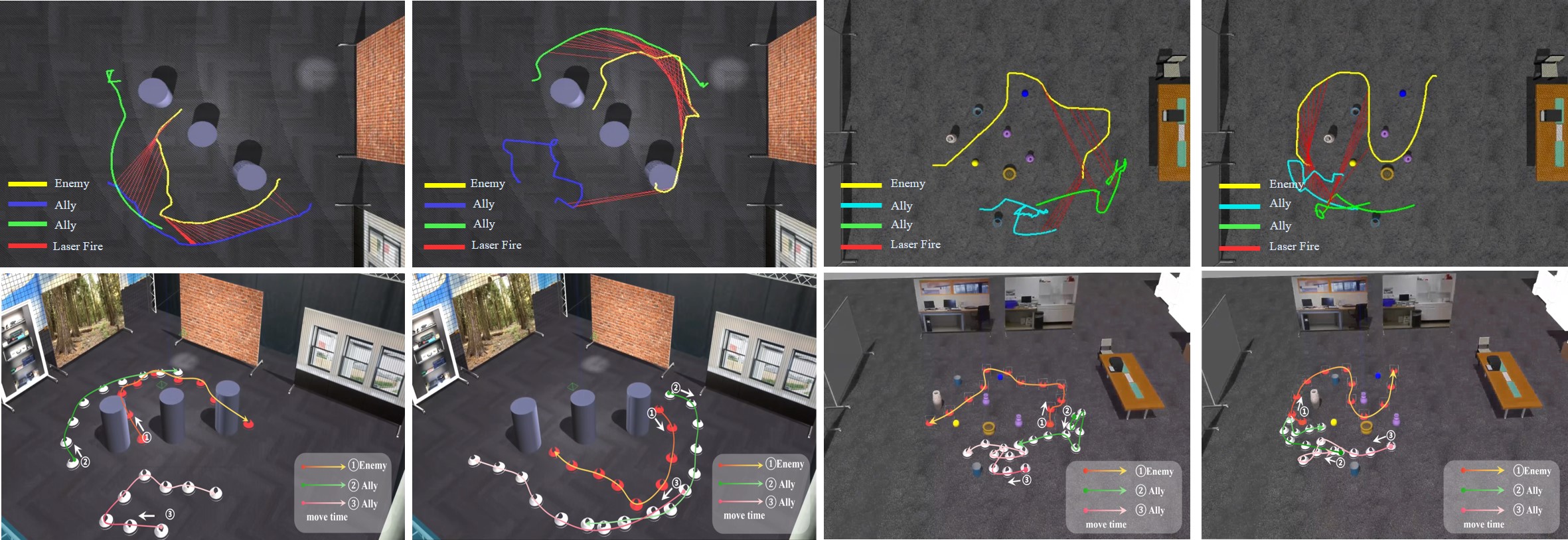}        
      \caption{The first row shows the trajectories of robots in the laser tag game in simulation. When the firing condition is met, an ally robot fires a laser beam toward the enemy, indicated by the red line. The second row shows the robot motions, with enemies in red and allies in white. Allies maneuver strategically to aim and attack the enemies, while the enemy maneuver to escape.}
      \label{figure_trajectory}
 \end{figure*}
Fig.~\ref{figure_detection} shows the onboard detection and depth images. We also use Full Gradient\cite{srinivas2019full} to analyse the attention area of the visuomotor policy. As shown, the activation areas on input images generally focus the enemy robot and environment obstacles. 
   \begin{figure}[thpb]
      \centering
      \includegraphics[scale=0.22]{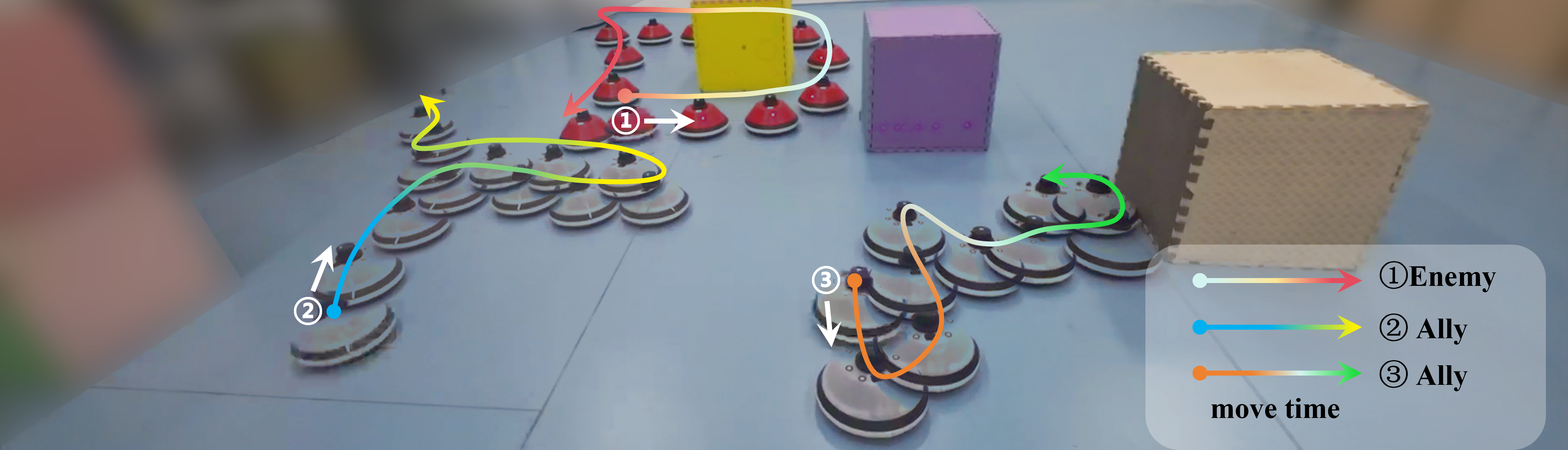}
      \caption{ Snapshots from the real-world experiment. }
      \label{figure_real_world}
 \end{figure}
       \begin{figure}[thpb]
      \centering
      \includegraphics[scale=0.078]{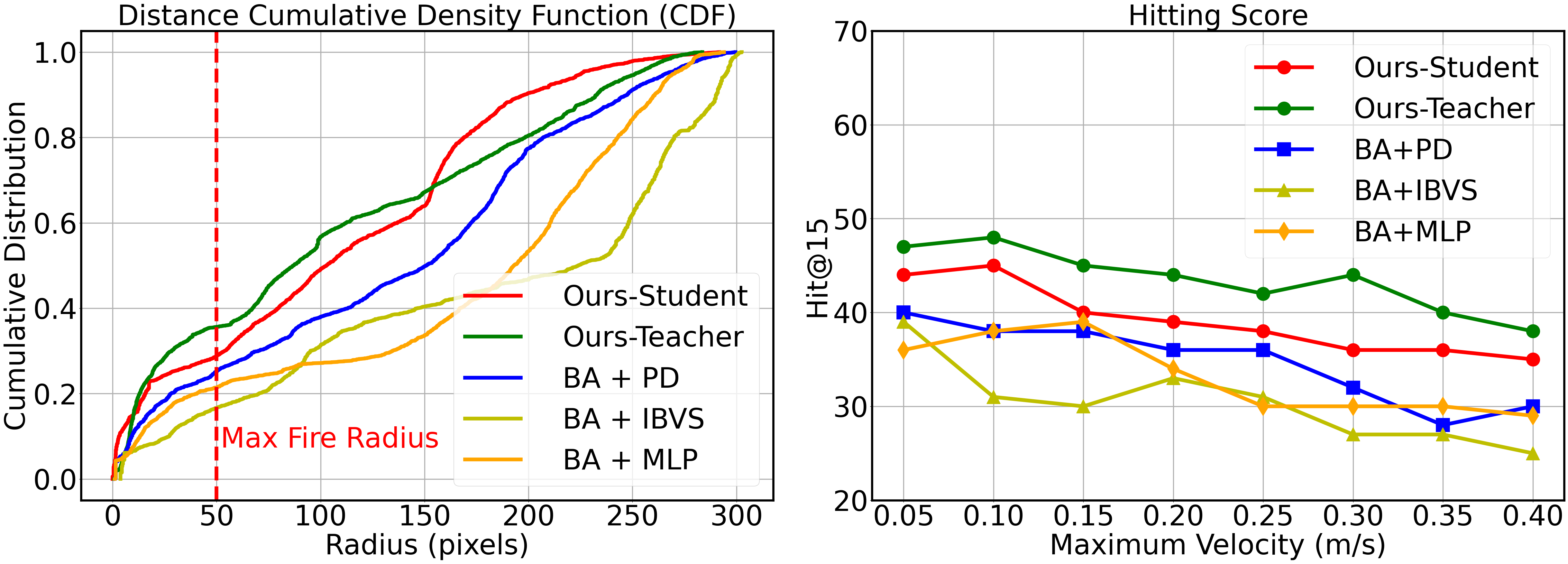}   
      \caption{ \textbf{Left}: Cumulative distribution function (CDF) of radial distances from the crosshair center. A steeper rise near the origin indicates that more shots are concentrated close to the crosshair, reflecting better laser aiming accuracy.
     \textbf{Right}: Hit@15 score under different maximum velocities. Both figures illustrate that ours have an improved laser hitting performance.}
      \label{hitting_vs_velocity}
   \end{figure}
\subsection{Comparison with Other Classic Methods}
\textbf{Baselines}. We compare our visuomotor policy against classic modular methods for shooting-game-style tasks. These classic approaches typically decouple the system into distinct modules for perception, state estimation, and control.
We evaluate several combinations of state estimation and control algorithms. For state estimation, we select baselines that rely solely on observations from monocular cameras, which is consistent with our problem setting (free of depth sensors). \textbf{Estimation Baseline 1} is the Bearing-Angle (BA) method \cite{Ning2024}, which estimates an enemy's relative position and velocity from bounding box detections. \textbf{Estimation Baseline 2} is Spatial-Temporal-Triangulation (STT) cooperative estimation\cite{zheng2025optimal}, which fuses image observations from multiple robots.
For the control policies, we consider three representative approaches. \textbf{Policy Baseline 1} is a classic Proportional-Derivative (PD) controller. \textbf{Policy Baseline 2} is an Image-Based Visual Servo (IBVS) \cite{chaumette2006visual}, which generates control commands from image-space errors and relative enemy depth. \textbf{Policy Baseline 3} is a neural controller \cite{bajcsy2024learning} that feeds estimated enemy positions and headings into an LSTM encoder followed by an MLP policy (referred to as MLP).
These estimation and control policies are integrated into complete pipelines. For the PD policy, we adopt GCOPTER \cite{WANG2022GCOPTER} as the trajectory planner, which generates waypoints to guide the ally to a desired shooting pose while avoiding collisions, using a pre-built voxel map. In contrast, IBVS and MLP directly output actions to aim at the enemy. Each method is evaluated over totally 50 runs, from random initial states in both simulation and real-world environments.

\begin{table}[h]
    \setlength\tabcolsep{3.4pt}
    \caption{Comparison with other methods. GT uses ground-truth states as input. Ours-Teacher refers to the state-based teacher policy, and Ours-Student to the vision-based student policy. Gray rows indicate methods relying on onboard sensing. }
    \begin{center}
    \begin{tabular}{l l c c c}
    \toprule
    \multicolumn{2}{c}{\textbf{Methods}} & \multirow{2}{*}{\textbf{Hit@15}$\uparrow$}  & \multirow{2}{*}{\textbf{Hit@30}$\uparrow$} & \multirow{2}{*}{\textbf{CR[\%]$\downarrow$}} \\
    \cmidrule(lr){1-2}
    \textbf{State Estimation} & \textbf{Control Policy} & & & \\
    \midrule
    GT                                  &   PD                                &  44.0 ± 1.8    & 20.0 ± 1.7       &  34.0               \\
    GT                                  &   MLP\cite{bajcsy2024learning} &  44.2 ± 3.0    & 20.2 ± 1.3       &  50.0               \\
    GT                                  &   IBVS\cite{chaumette2006visual}    &  35.2 ± 5.5    & 16.0 ± 5.1       &  68.0               \\
    GT                                  &  Ours-Teacher                       &  48.6 ± 1.7    & 22.1 ± 2.0       &  10.0               \\\rowcolor{gray!20} 
    
    BA\cite{Ning2024}                   &   PD                                &  33.0 ± 1.8    & 13.9 ± 2.4       &  42.0               \\\rowcolor{gray!20} 
    BA\cite{Ning2024}                   &   MLP\cite{bajcsy2024learning} &  32.5 ± 2.4    & 13.5 ± 2.1       &  54.0               \\\rowcolor{gray!20} 
    BA\cite{Ning2024}                   &  IBVS\cite{chaumette2006visual}     &  24.7 ± 4.9    & 12.2 ± 4.0       &  70.0               \\\rowcolor{gray!20} 
    
    STT\cite{zheng2025optimal}          &   PD                                &  20.2  ± 4.0   & 10.2 ± 4.8       &  46.0               \\\rowcolor{gray!20} 
    STT\cite{zheng2025optimal}          &   MLP\cite{bajcsy2024learning} &  13.2  ± 4.1   & 5.5  ± 3.1       &  60.0               \\\rowcolor{gray!20} 
    STT\cite{zheng2025optimal}          &   IBVS\cite{chaumette2006visual}    &  6.9   ± 3.9   & 2.6  ± 1.6       &  72.0               \\\rowcolor{gray!20} 
    
    N/A                                 & \textbf{Ours-Student}               &  \textbf{38.4 ± 2.2}    & \textbf{17.2 ± 2.8}       &  \textbf{36.0}               \\
    \bottomrule
    \end{tabular}
    \end{center}
    \label{tableCompare}
\end{table}

   \begin{figure}[thpb]
      \centering
      \includegraphics[scale=0.078]{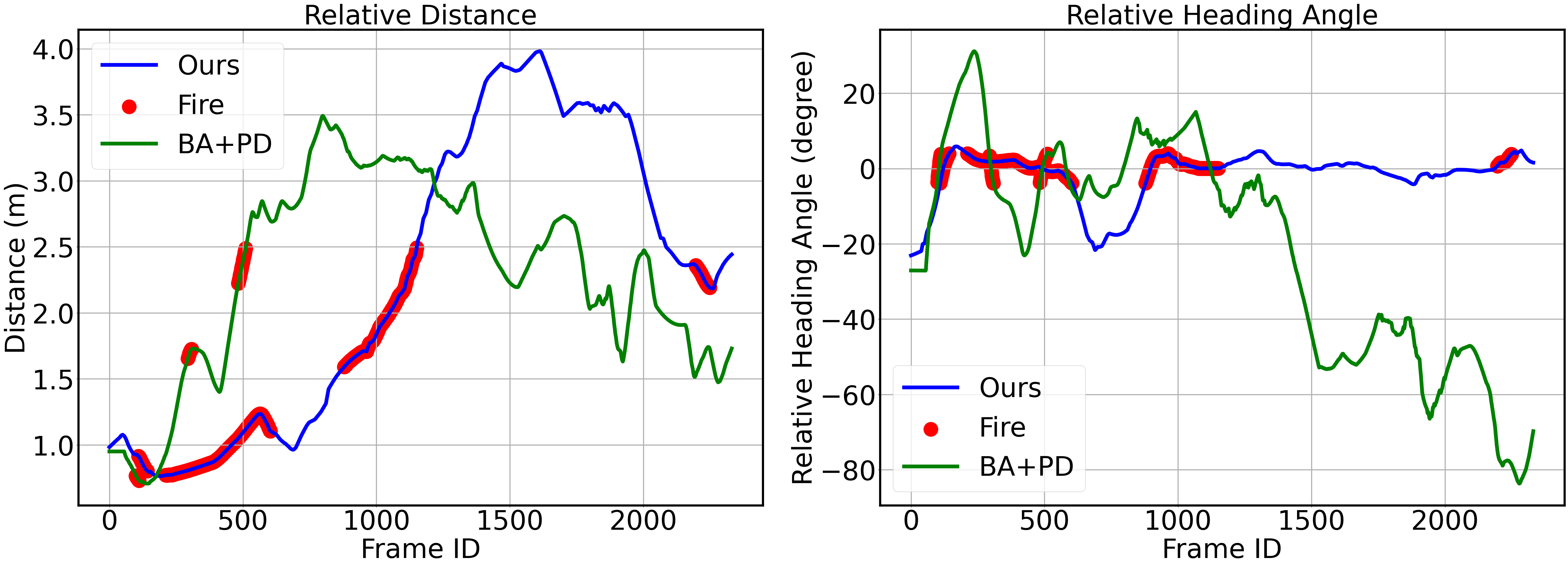}
      \caption{Relative distance and heading between the ally and enemy in one episode, with red segments marking when the enemy is within the shooting zone. The visuomotor policy yields more successful targeting frames than the PD policy.}
      \label{relative_distance_curve}
   \end{figure}

\textbf{Results}. Table~\ref{tableCompare} shows the comparison results.
For state estimation, BA achieves the best performance, with a position estimation error of 0.094 ± 0.021 m, while STT is less accurate (0.291 ± 0.072 m). The error in STT mainly arises from reliance on global camera poses, which are affected by odometry and calibration inaccuracies as well as internal estimator errors. Moreover, STT requires inter-robot communication to share camera poses.
For control, PD and MLP perform similarly under the same estimator and both surpass IBVS. Our visuomotor policy achieves the best onboard performance in laser-hitting and obstacle avoidance, improving by 16.1\% and 6\% respectively over the strongest baseline (BA + PD). Policies with access to ground-truth states perform better, confirming perception and state estimation as the dominant error sources. Finally, since IBVS and MLP inherently ignore obstacles, they suffer from higher collision rates.
Fig.~\ref{hitting_vs_velocity} (left) shows the cumulative distribution function (CDF) of the pixel-wise radial distance between the enemy bounding box center and the crosshair. 
The state-based teacher policy achieves the highest aiming accuracy, with 37\% of frames meeting the fire condition, while our student policy outperforms other onboard-sensing baselines with 28\%.
Fig.~\ref{hitting_vs_velocity} (right) reports the hitting score under different maximum velocity limits. 
For fairness, the maximum velocities of the ally and the enemy are set equal, and all methods are initialized from the same states. As shown in the figure, our visuomotor policy consistently achieves higher hit scores across different velocity limits.
Fig.~\ref{relative_distance_curve} illustrates the evolution of the relative distance and heading angles in a representative episode. 

\section{Conclusions}
In this paper, we study the problem of multi-robot laser
tag game with visuomotor policy. To address the challenges
of classic modular designs, such as estimation observability
limitation and depth-sensor-based-mapping, we propose an end-to-end learning
approach that directly maps raw image inputs to robot
actions. Experiments show that the visuomotor policy
outperforms classic modular methods regarding similar tasks
in terms of hitting accuracy and collision avoidance. We deploy the visuomotor policy on real-world multi-
robot system with limited computation resources, demonstrating the practicality of our approach. 

\textbf{Acknowledgment}. We would like to thank Shiliang Guo for his help in troubleshooting the experimental setup, and Zian Ning and Canlun Zheng for their insightful discussions.






\bibliographystyle{ieeetr}
\bibliography{ref3}
\end{document}